\documentclass{article}

\usepackage{microtype}
\usepackage{graphicx}
\usepackage{subcaption}
\usepackage{booktabs} 
\usepackage[table]{xcolor}
\usepackage{tabularx}
\usepackage{array}
\usepackage{graphicx}
\usepackage[utf8]{inputenc}

\usepackage{placeins}

\usepackage{hyperref}

\usepackage[preprint]{icml2026}

\usepackage{amsmath}
\usepackage{amssymb}
\usepackage{mathtools}
\usepackage{amsthm}
\usepackage{soul}

\usepackage[capitalize,noabbrev]{cleveref}

\theoremstyle{plain}

\theoremstyle{definition}

\theoremstyle{remark}

\usepackage[textsize=tiny]{todonotes}

\icmltitlerunning{Increasing Computation Resolves Conflicts in Vision Language Models}

\begin{document}

\twocolumn[
  \icmltitle{Increasing Computation Resolves Conflicts in Vision Language Models}



  \icmlsetsymbol{equal}{*}

  \begin{icmlauthorlist}
    \icmlauthor{Bingyang Wang}{equal,gt}
    \icmlauthor{Yijiang Li}{equal,ucsd}
    \icmlauthor{Yitong Qiao}{zju}
    \icmlauthor{Maijunxian Wang}{ucb}
    \icmlauthor{Tianwei Zhao}{jhu}
    \icmlauthor{Yucheng Sun}{eth}
    \icmlauthor{Binyue Deng}{gt}
    \icmlauthor{Hokin Deng}{cmu}
    \icmlauthor{Nuno Vasconcelos}{ucsd}
    \icmlauthor{Dezhi Luo}{um}
  \end{icmlauthorlist}

  \icmlaffiliation{gt}{Georgia Institute of Technology}
  \icmlaffiliation{ucsd}{University of California San Diego}
  \icmlaffiliation{zju}{Zhejiang University}
  \icmlaffiliation{ucb}{University of California Berkeley}
  \icmlaffiliation{jhu}{Johns Hopkins University}
  \icmlaffiliation{eth}{ETH Zurich}
  \icmlaffiliation{cmu}{Carnegie Mellon University}
  \icmlaffiliation{um}{University of Michigan}

  \icmlcorrespondingauthor{Bingyang Wang}{icy.bingyang.wang@alumni.emory.edu}
  \icmlcorrespondingauthor{Yijiang Li}{yijiangli@ucsd.edu}
  \icmlcorrespondingauthor{Dezhi Luo}{ihzedoul@umich.edu}

  \icmlkeywords{Machine Learning, ICML}

  \vskip 0.3in
]



\printAffiliationsAndNotice{}  

\begin{abstract}
  Cognitive control—the ability to coordinate competing information sources in pursuit of goals—is fundamental to intelligent behavior. We systematically investigate whether Vision Language Models (VLMs) exhibit cognitive control and how computational resources modulate conflict resolution. We construct a benchmark of 4,410 tasks across seven conflict paradigms (Stroop, Flanker, and five realistic variants) spanning multiple difficulty levels and visual complexities, testing 47 VLMs with rigorous experimental control. We find that VLMs exhibit robust congruency effects across all tasks, with larger models systematically resolving conflicts more effectively than smaller models. Critically, VLMs reproduce the fine-grained demand-resource relationship observed in human temporal dynamics: larger models drop below chance on incongruent high-conflict trials while smaller models fail to meaningfully engage and perform at chance, mirroring human behavior at short processing times and establishing parameter count as a proxy for conflict resolution capacity. These findings demonstrate that human-like cognitive control emerges from optimization dynamics in large-scale neural networks, suggesting that adaptive flexibility under conflict may naturally arise through scaling.

\end{abstract}

\section{Introduction}

Human behavior is distinguished by its flexibility and goal-directedness: we can pursue novel, underspecified tasks, adapt to changing contexts, and manage competing objectives and conflicting signals over time \citep{rasmussen1990role, botvinick2001conflict}. Central to these abilities is cognitive control—the mechanisms that dynamically coordinate thought and action in pursuit of goals \citep{egner2023principles, badre2024cognitive}. This adaptive control over competing signals makes cognitive control particularly critical in artificial general intelligence (AGI) \citep{anderson1983architecture, russin2020deep, lecun2022path} operating in the wild, where conflict arises from noisy sensors \citep{kendall2017uncertainties, hendrycks2019benchmarking}, miscommunication \citep{li2023evaluating, bitton2023visit}, and malicious adversarial attempts \citep{eykholt2018robust, kurakin2018adversarial}. 

Vision-language models (VLMs) \citep{liu2024visual, li2024llava, bai2025qwen2, abouelenin2025phi, openai2023gpt4, team2023gemini} have demonstrated unprecedented capabilities in integrating visual and textual information, achieving remarkable performance on perception and reasoning benchmarks \citep{li2024seed, fu2023mme}. However, it remains unclear whether these capabilities transfer to realistic scenarios where noisy and conflicting signals are prevalent \citep{yu2024natural, zhang2024mme}. This gap is particularly critical in safety-critical domains such as medical diagnosis, autonomous driving, and finance, where models must prioritize task-relevant information while suppressing misleading cues. Without robust mechanisms to coordinate competing information sources, VLMs may fail systematically under the ambiguity and conflict inherent in natural multimodal environments \citep{lee-etal-2025-vlind, wang2025text}.

\begin{figure*}[t]
    \centering
    \includegraphics[width=1\linewidth]{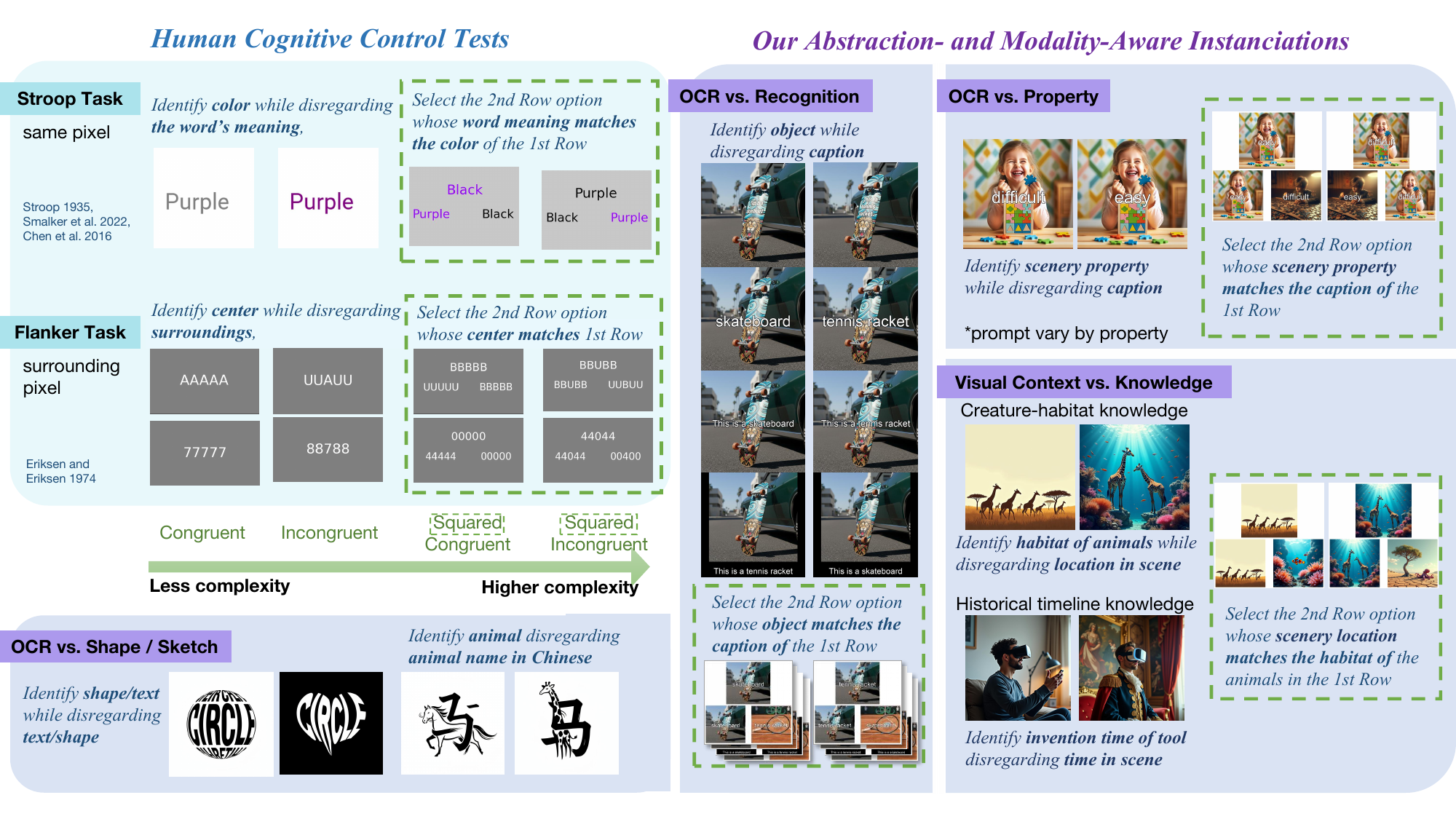}
    \caption{Cognitive Control Task Design Overview. \textbf{Blue panel} displays classic cognitive control paradigms from human studies, including the Stroop task (identifying color while disregarding word meaning) and Flanker task (identifying center stimulus while disregarding flanking distractors). \textbf{Purple panel} presents our multimodal adaptations that extend these paradigms to VLM-specific conflict spaces, systematically testing cognitive control across OCR, object recognition, visual properties, and knowledge reasoning. Tasks range from congruent to incongruent conditions with varying complexity levels, enabling measurement of the congruency effect and computational resource demands in conflict resolution.}
    \label{fig:main}

\vspace{-5mm}
    
\end{figure*}

This motivates a fundamental question: \textit{Can VLMs robustly coordinate conflicting signals in pursuit of goals?} Answering this requires systematic evaluation of cognitive control—the ability to flexibly configure behavior to prioritize goal-relevant information under interference. While cognitive control has been extensively studied in humans, its manifestation in VLMs remains largely unexplored. We address this by constructing a comprehensive benchmark of 4,410 tasks spanning seven conflict paradigms, tested across 47 VLMs. Our benchmark systematically adapts classic cognitive control tasks to multimodal settings while extending them to photo-realistic scenarios, programmatically generating controlled stimuli that vary perceptual factors across conflict variants, including high-difficulty "squared" versions that introduce multiple simultaneous conflicts.

Our findings reveal four key results. First, VLMs exhibit robust congruency effects: accuracy consistently decreases on incongruent relative to congruent trials, demonstrating genuine susceptibility to cognitive conflict. Second, increased model scale systematically enhances conflict resolution, with larger models achieving comparable performance across conditions while smaller models show pronounced congruency effects, establishing parameter count as a proxy for conflict resolution capacity. Third, VLMs reproduce the fine-grained demand-resource relationship observed in human temporal dynamics: larger models drop below chance on incongruent high-conflict trials while smaller models perform at chance, and performance as a function of log parameter count approximates human forced-response time-course functions. Fourth, this characteristic below-chance ``dip" on incongruent trials guards against spurious correlations—demonstrating that models exhibit genuine interference rather than monotonic improvement with scale, a signature absent in previous connectionist models of cognitive control. These findings demonstrate that human-like cognitive control have emerged in VLMs.


\vspace{-3mm}

\section{Related Works}


\subsection{Cognitive Control}

Cognitive control enables the regulation of thought and action in pursuit of goals, particularly in situations involving conflict, ambiguity, or distraction \citep{diamond2013executive, hammond1972cognitive}. Control operates by coordinating information flow through a limited-capacity system \citep{luck1997capacity, awh2007visual}, selecting which sensory cues, contextual signals, and goals are actively maintained in working memory and determining when these representations should be updated or suppressed \citep{smith1999storage, engle2002working}. When overlapping sources of information are not well separated—such as when conflicting features map to different responses—interference arises and performance suffers \citep{cohen1990control}. Control mitigates this by modulating representational strength and prioritization \citep{abrahamse2016grounding, verguts2009adaptation}, allowing goal-relevant information to guide behavior efficiently under constraints of time, uncertainty, and limited resources \citep{miller2001integrative, duncan2013structure, koechlin2003architecture}.

Conflict tasks are widely used to probe these mechanisms, with behavioral measures such as reaction time and error rates serving as proxies for internal computational demands \citep{braver2012variable, matsumoto2004conflict}. These paradigms manipulate congruency: congruent trials align task-relevant and task-irrelevant features, whereas incongruent trials pit them against each other, requiring inhibition of a salient but misleading cue \citep{macleod1991half, egner2005cognitive}. Incongruent trials reliably produce slower processing times (PTs) and lower accuracy than congruent trials—a pattern known as the \textbf{congruency effect}, which serves as the key behavioral marker of interference resolution \citep{macleod1991half, matsumoto2004conflict, notebaert2006top, verbruggen2006stimulus}. The magnitude of this effect reflects the degree of control exerted to suppress misleading cues and prioritize task-relevant information \citep{braver2012variable, egner2005cognitive}.

Human psychophysics has revealed temporal dynamics in conflict processing: key signatures of the congruency effect emerge at shorter PTs, while longer PTs allow systematic increases in accuracy (for further details, see Appendix \ref{app:processing_time}). Consistently, tasks with higher levels of conflict demand greater computational resources, as participants typically allocate more PT to resolve high-conflict trials compared to low-conflict ones. Together, these findings reflect a demand-resource relationship in which cognitive control scales with available computational resources. Developing unified frameworks that account for the nature of cognitive control is a central aim of cognitive science \citep{salvucci2011toward}, and an important guide for understanding general-purpose intelligence in both biological and artificial systems \citep{russin2020deep}.

\subsection{Benchmarking Vision-language Models}

No work has directly evaluated cognitive control in vision-language models (VLMs) \citep{team2023gemini, openai2023gpt4}, although these systems achieve near-human performance on perceptual and reasoning tasks that plausibly scaffold control, such as spatial reasoning  \citep{chen2024spatialvlm, cai2024spatialbot}, OCR \citep{mathew2021docvqa, li2023trocr}, scene understanding \citep{li2024seed, liu2024visual, fu2023mme, liu2023hidden, fu2024scene} and cognitive abilities \citep{li2025core}. Their progress is largely attributed to web-scale multimodal pretraining and improved cross-modal alignment by adapting LLMs \citep{li2024seed, fu2023mme, wu2024v, xu2024llava, shao2024visual}. Existing studies of “conflict” in VLMs mostly examine modality disagreements (e.g., reliance on language priors over visual evidence), cross-modal inference, or inconsistencies between local and global visual signals, often finding degraded reasoning when cues conflict \citep{lee-etal-2025-vlind, wang2025text, li2023multi, cho2025ignorance, pandey2022crossmodal, Chen2024LION}. More recently, dedicated benchmarks have begun to study modality-\citep{zhang2025robust, pandey2022crossmodal} and knowledge-level \cite{jia2025benchmarkingmultimodalknowledgeconflict} conflicts, revealing persistent failures to maintain coherent decisions when evidence is competing or evolves over context. By contrast, our work introduces the first unified framework grounded in classic human conflict paradigms to systematically evaluate the congruency effect and ability to resolve such conflict (i.e., cognitive control) in VLMs across OCR, object recognition, visual attributes, and knowledge domains.


\section{Experiment Design}

\subsection{Cognitive Framework}
Cognitive control is commonly studied using conflict tasks that require integrating multiple information sources while suppressing interference \citep{fan2003cognitive, veen2006conflict}. These paradigms vary whether task-relevant and irrelevant features align (congruent) or conflict (incongruent). Canonical tasks such as Stroop and Flanker \citep{stroop1935studies, eriksen1974effects} operationalize conflict resolution through the congruency effect—the reliable performance cost on incongruent relative to congruent trials that indexes the computational demands of interference control \citep{badre2024cognitive}. Abstractly, a conflict task presents two compliant or conflicting signals embedded within an image, each corresponding to a distinct objective, with instructions designating one signal as goal-relevant and the other irrelevant. For instance, in the Stroop task \citep{friedman2022role}, a word's semantic content (e.g., ``RED'') and its ink color (e.g., blue) constitute two entangled signals. On incongruent trials, the model is required to prioritize the task-relevant signal and suppress the prepotent but incorrect alternative. This captures the fundamental computational challenge: selecting and amplifying goal-relevant information while suppressing strong competing signals.
\subsection{Scaling Up Conflict Tasks}

\textbf{Scaling up via Visual Programming.}

We construct multimodal variants of classic cognitive-control tasks via visual programming. In Stroop \citep{stroop1935studies}, conflict arises between a word’s meaning and its ink color (e.g., ``Purple'' printed in the color of orange). In Flanker \citep{eriksen1974effects}, models identify a central target (e.g., an arrow or emoji) while ignoring surrounding icons (distractors). To avoid performance saturation, we also implement squared version of Stroop and Flanker tasks \citep{burgoyne2023nature}, as illustrated in Figure~\ref{fig:classic_conflict}, which add a second, independent conflict (e.g., color-position mappings).

To scale up the classic Stroop and Flanker for more rigorous and comprehensive evaluation, develop a generation pipeline that programmatically varies both stimulus content and perceptual factors, producing hundreds of thousands of images. For Stroop, we enumerate combinations of 15 colors and their corresponding words. For Flanker, targets and distractors are drawn from a diverse symbol set, including arrows, numbers, letters, and emojis. Across tasks, we randomize font family (e.g., sans-serif, serif, monospace), font and icon size, spatial position, and background color. This procedure scales classic conflict tasks from a small set of templates to a large, balanced corpus spanning the full combinatorial space of congruent and incongruent conditions. Collectively, these tasks probe rapid interference resolution under binary forced-choice decisions.

\textbf{From Synthetic to Photo-realistic.}
Our goal is to evaluate VLMs’ ability to resolve conflict while pursuing task goals in realistic settings. Therefore, we move beyond synthetic, visually programmed stimuli and curate photo-realistic images, tasks, and questions that reflect real-world visual complexity and practical use cases. Drawing on the classic conflict paradigms described above, we design a suite of naturalistic tasks that embed controlled conflicts within photo-realistic scenes.

\textbf{\ul{Stroop-Object (overlay text vs. pixel object)}.}
Stroop-Object (overlay text vs. pixel object) tasks introduce conflict between the depicted object and overlaid text that names an object category. This task evaluates whether models can resolve competition between OCR-derived text semantics and visual evidence, in pursuit of the task goal of OCR or object recognition.
We use photo-realistic images (e.g., from COCO) and overlay potentially conflicting labels (e.g., a tennis racket labeled ``skateboard''). Text is integrated at varying levels of realism, from simple high-contrast overlays (see Figure~\ref{fig:coco_natural_conflict}) to naturalistic embeddings on the object surface (Figure~\ref{fig:object_conflict}). This design tests whether models remain grounded in pixel-level visual content rather than defaulting to a ``reading" shortcut in text-rich scenes.

\textbf{\ul{Stroop-Attribute (overlay text vs. pixel properties).} }
We incorporate fine-grained visual understanding by making pixel-level properties the task-relevant signal (e.g., geometric shape, scene atmosphere such as modern vs.\ old-fashioned, and human affect such as happy vs.\ sad; see Figure~\ref{fig:property_conflict}). In the Stroop-Attribute task, each image contains an overlaid descriptor that is either congruent or incongruent with the underlying attribute, requiring the model to resolve between text semantics and visual properties in pursuit of different goals such as OCR and attribute recognition.  

A noteworthy subset is the shape conflict task using calligrams: the semantic label is rendered in a mismatched visual form (e.g., the word “CIRCLE” shaped as a heart-like contour), explicitly decoupling lexical meaning from geometric appearance. We further introduce symbolic conflict (Pictographic vs.\ Hieroglyphs) by pairing a Chinese character with an incompatible pictorial cue (e.g., the character for “horse” combined with a drawing of another animal). Because the distractor is itself a meaningful visual symbol—not merely alphabetic text—these stimuli provide a unique test of conflict resolution in the pixel representation.


 

\textbf{\ul{Visual Context and Knowledge Conflicts (Visual Context vs. Knowledge Prior).} }
We evaluate models’ ability to resolve conflicts between parametric knowledge and immediate visual context using \textbf{Visual Context and Knowledge Conflicts} tasks. For example, \textit{Creature--Habitat} (Figure~\ref{fig:prior_knowledge}) depicts an animal in an atypical environment (e.g., a giraffe shown in an ocean scene) while querying its natural habitat. Success in incongruent cases requires suppressing task-irrelevant visual context and retrieving the appropriate parametric knowledge acquired during training.

To systematically evaluate cognitive control in VLMs, we construct a benchmark of 4,410 tasks, comprising 3,616 unique images and 4,410 question-answer pairings across seven categories. These categories include four classic conflict tasks (Stroop, Squared Stroop, Flanker, and Squared Flanker) and three realistic conflict tasks (Stroop-Object, Stroop-Attribute, and Visual Context \& Knowledge Conflicts). This scale enables rigorous measurement of conflict resolution across a broad range of controlled configurations (e.g., text size and placement), supporting robust and generalizable conclusions. Moreover, the benchmark covers multiple levels of abstraction and difficulty, allowing us to test whether observed effects persist from synthetic settings to more realistic, practically relevant scenarios.

\textbf{Implementation \& Scaling}

Our implementation follows a systematic three-stage pipeline that generates controlled conflict stimuli across multiple abstraction levels. For stroop-object conflicts, we use a unified process based on Gemini-2.5-flash-image-preview (Appendix~\ref{appen_pipeline_for_aigen}). As illustrated in Figure~\ref{fig:nanobanana examples}, each instance begins with a base image of a target category, then introduces conflicting text (e.g., This is a truck'' over a car image), and finally produces a matched version by replacing the text with a congruent label (e.g., This is a car'') while preserving the same style, placement, and typography.
This design ensures that congruent and incongruent conditions share identical visual content, with differences attributable solely to the textual manipulation rather than uncontrolled image variation. The same mechanism supports conflicts at different abstraction levels, including object misidentification, attribute misattribution, and contradictions of prior knowledge. For squared comparison tasks, we extend this pipeline by generating reversed base images and creating systematic pairings between triplets (base + conflict + matching). Each conflict-matching pair from the same base image is spatially arranged in a 2×2 grid format, enabling direct visual comparison while maintaining the controlled conflict manipulation (see Figure \ref{fig:main}).




 \subsection{Quality Control}
All annotated items underwent a rigorous human review to ensure quality and consistency. Each item was first inspected by an annotator using four criteria: (1) pair comparability, verifying that the target and option images were visually and semantically matched in difficulty; (2) diffusion-generation errors, screening for artifacts or implausible outputs from generative models; (3) question clarity, ensuring prompts were unambiguous and grammatically natural; and (4) task-specific compliance, confirming alignment with the intended conflict type.
Each item then underwent an independent blind review: a second reviewer answered the question without access to the original annotation, providing a check on both comprehensibility and label correctness. Annotator–reviewer disagreements were flagged and adjudicated until consensus, yielding reliable final annotations. The post-review counts are as follows:
\begin{table}[h]
\centering
\footnotesize

\caption{Distribution of tasks in our benchmark.}
\label{tab:taskcount}

\setlength{\tabcolsep}{6pt}
\renewcommand{\arraystretch}{1.15}

\begin{tabularx}{\columnwidth}{@{}
  >{\raggedright\arraybackslash\hsize=0.80\hsize}X r
  >{\raggedright\arraybackslash\hsize=1.20\hsize}X r
@{}}
\toprule
\multicolumn{2}{@{}l}{\textbf{Classic Conflict Task}} &
\multicolumn{2}{l@{}}{\textbf{Realistic Conflict Task}} \\
\midrule
Stroop   & 488 & \mbox{Stroop-Object}            & 3104 \\
Squared Stroop  & 640 & \mbox{Stroop-Attribute}               & 1170 \\
Flanker  & 324 & \mbox{Visual Context and}  & 136 \\
Squared Flanker & 1280 & \mbox{Knowledge Conflicts} \\
\addlinespace[2pt]
\midrule
\textbf{Total} &  &  & \textbf{4410} \\
\bottomrule
\end{tabularx}

\end{table}

\vspace{-8mm}

\section{Experiment Results}

\subsection{Hypotheses}

By applying conflict tasks to study cognitive control in VLMs, we seek to investigate \textbf{(1) the extent to which models are subjected to cognitive conflict}, assessed by comparing accuracy on congruent versus incongruent trials (the congruency effect), and \textbf{(2) whether models can leverage computational resources to resolve conflict}, examined by testing whether models with more parameters exhibit smaller congruency effects. A central principle in machine learning holds that scaling up model size yields systematic improvements in reasoning and generalization capabilities \citep{sutton2019bitter, kaplan2020scaling}, making parameter count a natural index for available computational resources. By calculating model accuracy across conditions as a function of parameter size, we obtain measurements directly comparable to human performance as a function of PT.

To examine whether \textbf{(3) models exhibit fine-grained demand-resource relationships in cognitive control} as observed in human temporal dynamics \citep{zhang2024temporal, lee2025forced}, we vary conflict difficulty within each paradigm. We predict that larger models will exhibit stronger congruency effects under high-conflict conditions than under low-conflict conditions, while smaller models will fail to meaningfully engage with sufficiently difficult high-conflict scenarios, performing at chance. This graded design also helps \textbf{(4) guard against spurious correlations}: connectionist models may appear to resolve conflicts when they actually rely on superficial shortcuts, producing monotonic performance increases with scale that are unable to disambiguate between the two. Crucially, if larger models drop below chance accuracy on incongruent trials of sufficiently high-conflict tasks while smaller models perform at chance, this dissociation would indicate genuine conflict processing rather than shortcut exploitation.

\subsection{Inference and Evaluation}
We evaluated a diverse set of VLMs spanning a wide range of architectures, parameter sizes, and training paradigms. In total, 47 different VLMs are tested. The open-source models ranging from 1B to 110B parameters enables comprehensive analysis across scales. Proprietary models were evaluated via API calls from standard personal computers. For open-source models, we ran inference on a cluster with 8× NVIDIA A100 80GB GPUs: models $<$13B typically used one GPU, 13B–32B used two, 32B–70B used four, and $>$70B used all eight. We followed the official inference codebases released by model developers to maintain reproducibility To standardize multimodal input handling and response parsing. To ensure fairness, a unified evaluation toolkit is adopted to validate and aggregate outputs across heterogeneous model interfaces.


\begin{figure}[h]
    \centering
    \includegraphics[width=1.0\linewidth]{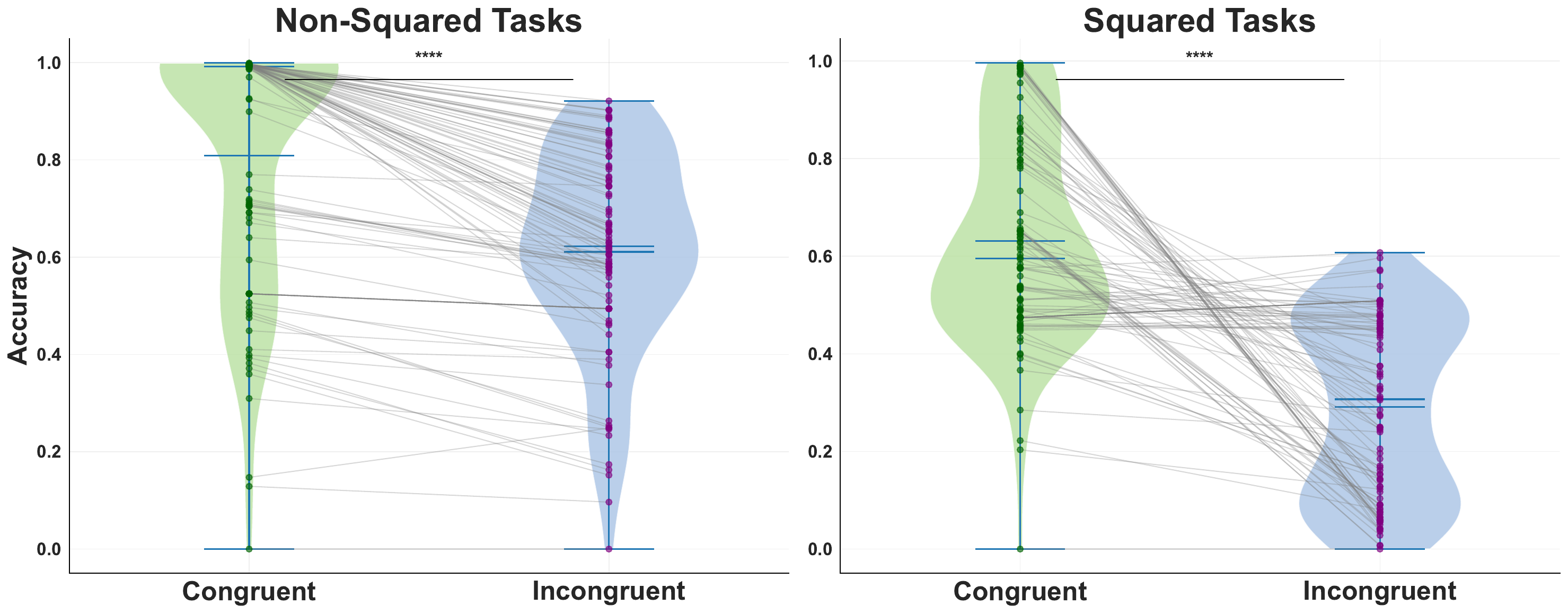}
    \caption{Distribution of accuracies under congruent (blue) and incongruent (orange) conditions. 
    Each connected pair of points corresponds to the same model evaluated under both conditions. 
    The consistent downward trend highlights a clear performance drop in incongruent trials, demonstrating consistent congruency effects across VLMs. See Table \ref{tab:combined-accuracies} for breakdown across task types.}
    \label{fig:main}
\end{figure}

\vspace{-5mm}

\subsection{Congruency Effect}

\textbf{Main Results.}
We observe a robust Congruency Effect across all tested visual-language models (VLMs): accuracy is substantially lower for incongruent than congruent image–text pairs, indicating that semantic interference reliably disrupts multimodal reasoning. In non-squared tasks, mean accuracy drops from 0.852 (congruent) to 0.567 (incongruent), a mean paired difference of 0.285 ($t=13.981$, $p=3.6\times10^{-25}$). In squared tasks, which impose additional spatial/compositional demands, the drop is larger—from 0.634 to 0.297 (mean difference 0.337; $t=10.023$, $p=9.8\times10^{-17}$). Thus, congruency effects not only persist but amplify when models must integrate multiple perceptual or textual dimensions. Consistent with this aggregate pattern, Figure~\ref{fig:main} shows that nearly all models (connected points) exhibit decreased accuracy from the congruent to incongruent condition, underscoring the effect’s robustness across architectures.





To further validate these results, we examine performance by task configuration (Fig.~\ref{fig:non-squared} and Fig.~\ref{fig:squared}). The same pattern holds throughout: accuracy is consistently and significantly lower in incongruent than congruent conditions. In non-squared tasks (Fig.~\ref{fig:non-squared}), the Congruency Effect is largest for Flanker-style and property-level conflicts, where perceptual–semantic interference directly competes with local feature recognition (mean differences $\approx 0.43$--$0.45$, $p<10^{-18}$). Object-level manipulations (e.g., word or sentence overlays) also produce reliable declines (0.20--0.24), indicating that even superficial textual incongruence disrupts object classification. In squared tasks (Fig.~\ref{fig:squared}), which require joint spatial/compositional and textual reasoning, the effect strengthens further: the largest gaps occur for object- and property-level conflicts (0.40--0.43), consistent with heightened vulnerability when visual and linguistic cues are simultaneously misleading or spatially entangled.
Together, these results illustrates a prominent Congruency Effect consistently across conflict types and task complexity.

\textbf{Modality Preference.} Our analysis of reversed-modality task pairs reveals a systematic but task-dependent modality preference across vision-language models. In Stroop-object tasks, where models must resolve conflicts between OCR and object recognition modalities, we observe a pronounced bias toward text-based processing: 35 out of 47 models (74.5\%) exhibit a strong preference for OCR targets (difference > 0.05), achieving a mean accuracy of 0.820 when answering text questions compared to 0.721 when answering object questions, resulting in a mean difference of 0.100 (SD = 0.255). This preference persists across model sizes, as indicated by the moderate positive correlation (r = 0.611) between performance on both tasks, which suggests that while models performing well on one task tend to perform well on the other, there is a consistent upward shift favoring OCR. In contrast, Stroop tasks involving OCR and color modalities reveal a more balanced distribution: 37 out of 43 models (86.0\%) demonstrate balanced performance (|difference| $\leq$ 0.05), with only 4 models (9.3\%) showing strong OCR preference. The mean difference in Stroop tasks (0.084, SD = 0.308) is smaller than in Stroop-object tasks, and the lower correlation (r = 0.243) indicates more independent processing between OCR and color modalities. The observed distribution patterns suggest that modality preference is not uniform across conflict types but rather emerges more strongly in cognitively demanding visual tasks. The unimodal distribution shifted toward OCR preference in Stroop-object tasks indicates a systematic architectural or training bias favoring text processing, likely stemming from extensive training on text-rich data and the relative ease of extracting textual features compared to complex object classification. This bias appears to be a fundamental characteristic of current VLM architectures rather than a task-specific artifact, as evidenced by its consistency across different model sizes and architectures.


\begin{figure}[htbp]
    \centering
    \begin{subfigure}[b]{1.0\linewidth}
        \centering
        \includegraphics[width=1.0\linewidth]{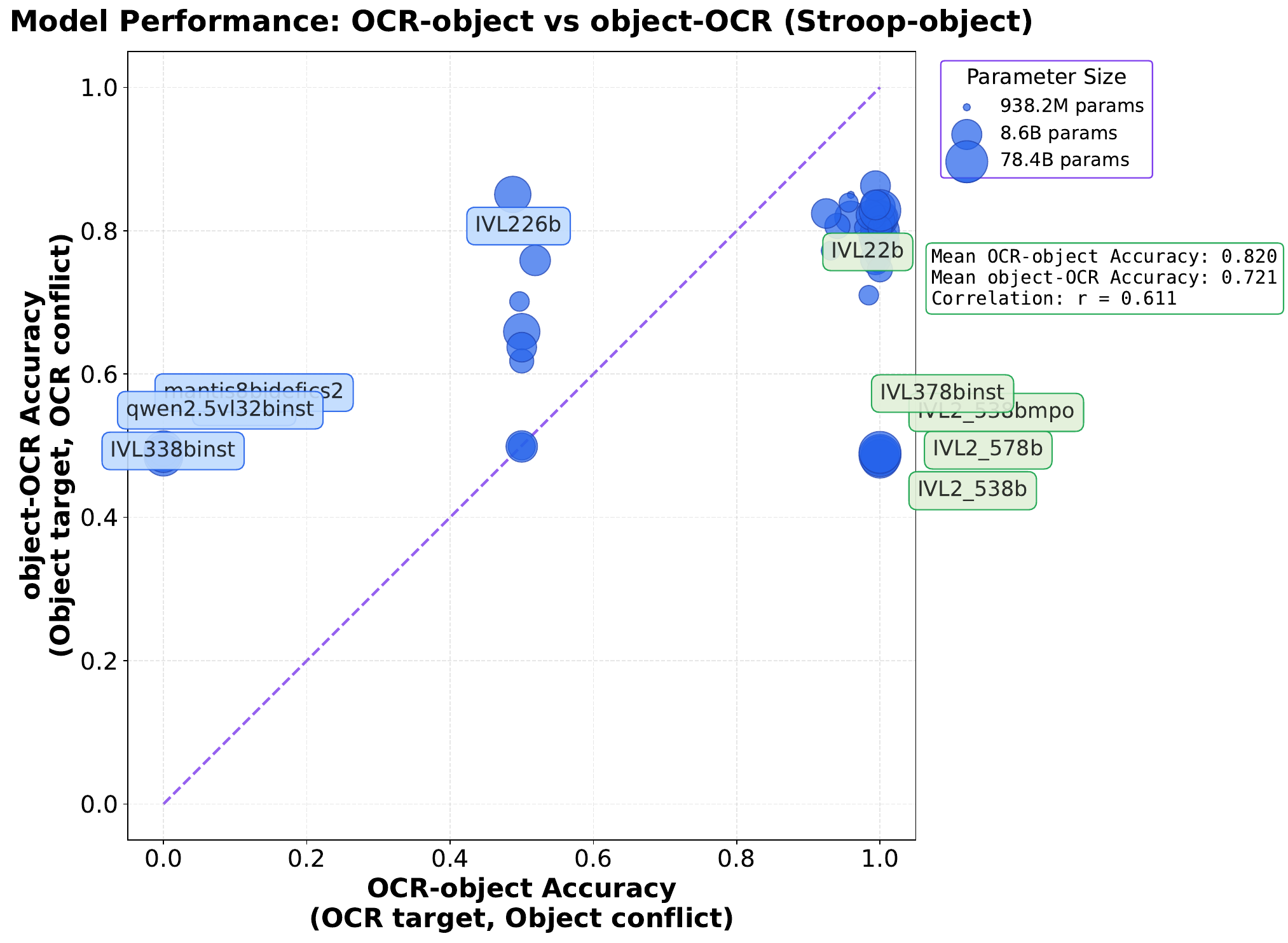}
        \label{fig:object_vs_ocr}
    \end{subfigure}
    \begin{subfigure}[b]{1.0\linewidth}
        \centering
        \includegraphics[width=1.0\linewidth]{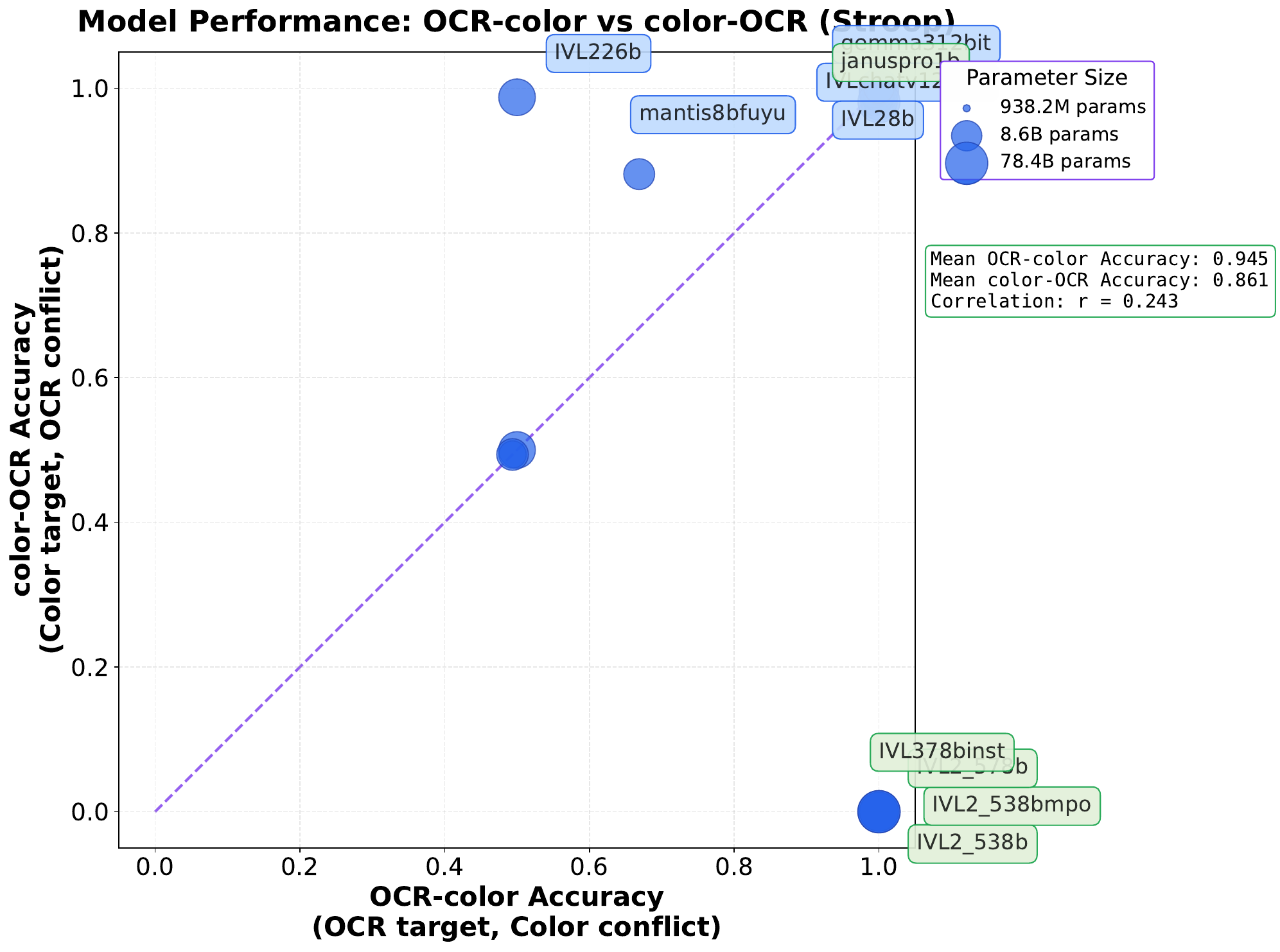}
        \label{fig:color_vs_ocr}
    \end{subfigure}
    \caption{Comparison of model performance under competing modalities. The figures illustrate how models perform when tasked to identify different attributes (Object/Color vs. Text) from the exact same visual input.}
    \label{fig:stroop_comparison_combined}
\end{figure}

However, the contrasting pattern in Stroop tasks—where the majority of models achieve balanced performance—suggests that simpler visual modalities (color recognition) may be more comparable to OCR in terms of processing demands, allowing models to flexibly attend to either modality as required. This task-dependent variation in modality preference has important implications for cognitive control evaluation: benchmarks must account for inherent modality biases by testing both directions of target-conflict reversal, and models demonstrating balanced performance across modalities may represent superior cognitive control capabilities, as they can suppress task-irrelevant information regardless of its modality. The finding that scaling alone does not eliminate modality preference further suggests that future improvements in balanced modality processing will require explicit architectural modifications or training objectives rather than simply increasing model capacity.



\subsection{Increasing Computation Resolves Conflicts}
\label{sec:model_performance_compute}

\begin{figure}[t]
    \centering
    \includegraphics[width=1.0\linewidth]{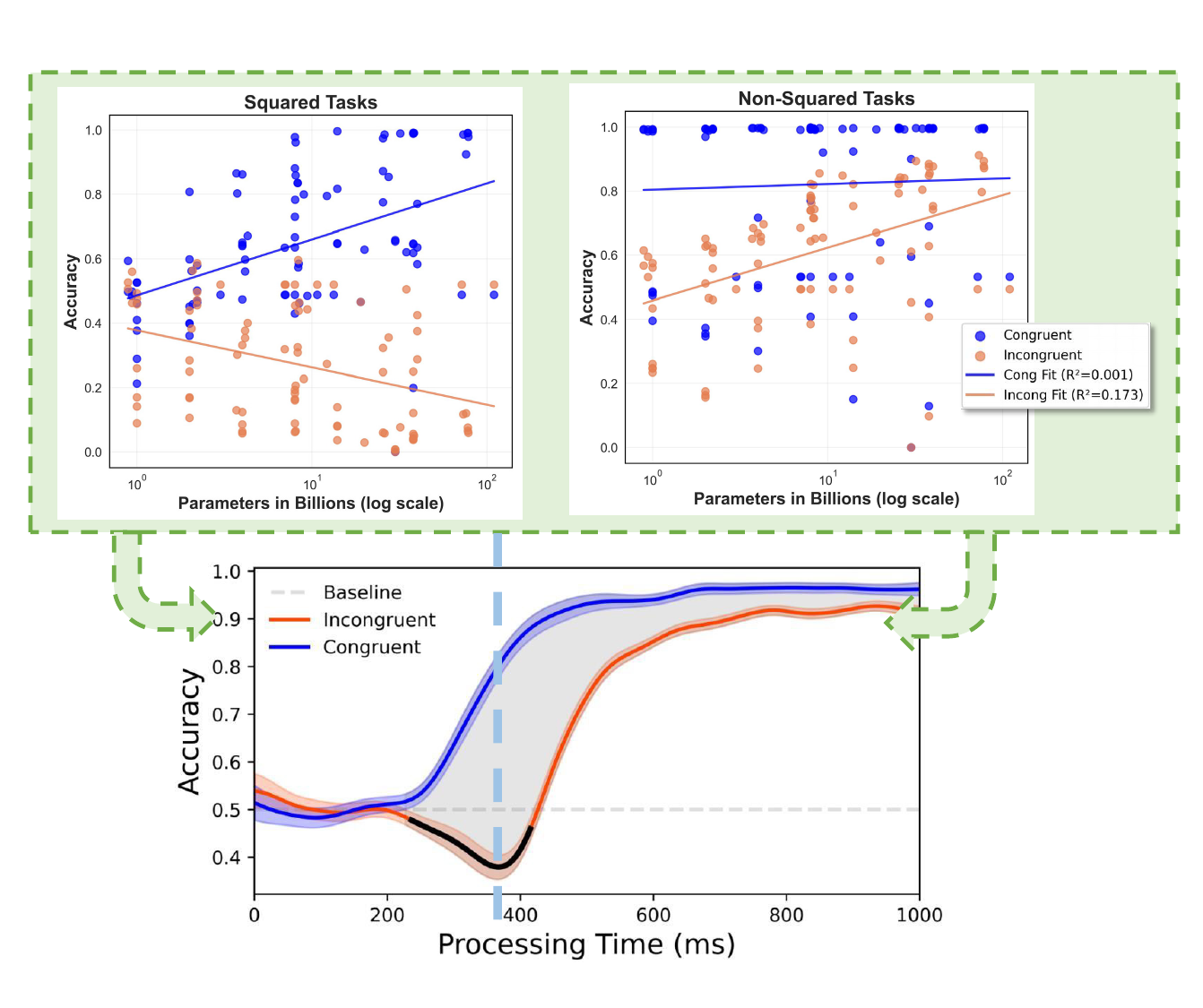}
    \caption{Model performance as a function of log parameter count approximates a cross-sectional processing curve analogous in form to the time-course functions observed in human forced-response paradigms \citep{lee2025forced}.}
    \label{fig:FR}
    \vspace{-7mm}
\end{figure}

We further analyze model performance in relation to computational resources by comparing models of varying sizes on both standard (low-conflict) and squared (high-conflict) trials. In Figure \ref{fig:FR}, we analyze performance on classic conflict tasks as a representative example. We begin with the standard (non-squared) tasks (top right panel), where a clear congruency effect emerges among smaller models: accuracy on incongruent trials is substantially lower than on congruent trials. In contrast, larger models exhibit comparable accuracy across both conditions, suggesting that increased computational resources, indexed by parameter count, enable conflict resolution. This trend parallels the human performance curve (bottom panel), where longer processing times (PT $>$ 400~ms) substantially reduce the congruency effect, reflecting successful conflict resolution with additional computational resources.

To assess whether this improvement in model performance reflects genuine conflict processing rather than reliance on superficial shortcuts, we turn to the squared (high-conflict) tasks (top left panel). In these tasks, smaller models perform at chance on both congruent and incongruent trials, indicating an inability to meaningfully engage with the task. Larger models, however, show above-chance performance on congruent trials but a marked drop to below-chance accuracy on incongruent trials. This dissociation suggests that increased model capacity allows for deeper processing of conflicting inputs, thereby exposing models to interference—an indicator of genuine conflict sensitivity. This pattern closely mirrors human behavior under short PT (approximately 0--400~ms), where participants transition from being unable to engage with the task (0--200~ms) to exhibiting strong congruency effects once minimal processing begins but before conflict is fully resolved.

This shift in model behavior reflects the expected interaction between task difficulty and computational resources: performance on high-conflict trials given sufficient resources is analogous to performance on low-conflict trials under limited resources. Taken together, the results show that performance as a function of log parameter count approximates a cross-sectional slice of the time-course functions observed in human forced-response paradigms (bottom panel), demonstrating a fine-grained demand-resource relationship in cognitive control. This reveals a structural correspondence between human and model conflict resolution, where PT in humans and parameter count in VLMs serve as functionally analogous resource variables. The characteristic "dip" in accuracy on incongruent trials during early human processing is replicated in larger models under high-conflict conditions, providing strong evidence that these models are genuinely engaged in conflict processing. Critically, this pattern indicates that performance gains with increased scale reflect an enhanced capacity for cognitive control, rather than reliance on superficial shortcuts.

\begin{figure*}[t]
    \centering
    \includegraphics[width=0.8\linewidth]{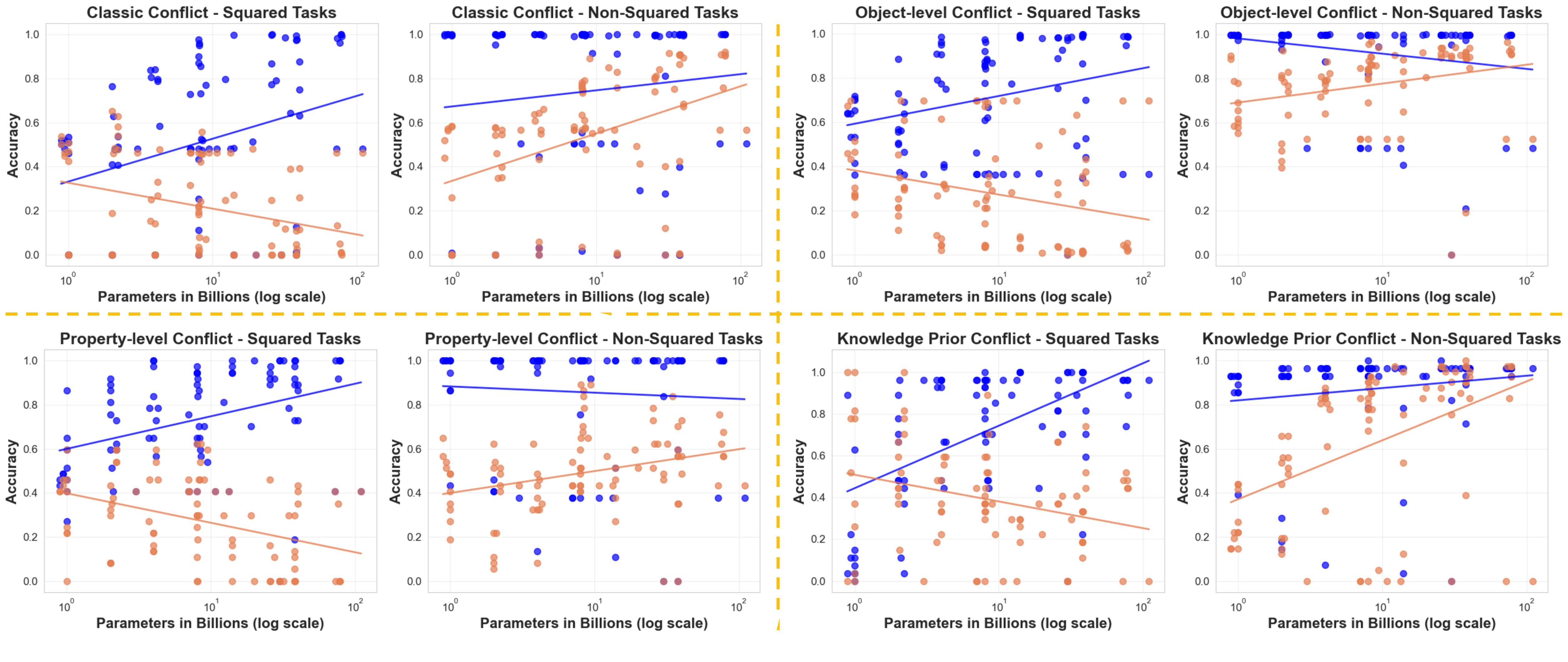}
    \caption{Model performance in relation to computational resources across all conflict types.}
    \label{fig:all_types}

    \vspace{-5mm}
\end{figure*}

We further show that this pattern is replicated across all four conflict types (Figure \ref{fig:all_types}), suggesting that the relationship between cognitive control and computational resources is robust and generalizes across diverse task structures.

\vspace{-3mm}

\section{Discussions}
Our results provide strong evidence that VLMs exhibit systematic signatures of cognitive control that parallel human conflict processing. We observed robust congruency effects across all tested models and task types: accuracy consistently decreased on incongruent relative to congruent trials, demonstrating that models are genuinely subject to cognitive conflict \textbf{(Hypothesis 1)}. Critically, increased computational resources enable conflict resolution \textbf{(Hypothesis 2)}: while smaller models showed pronounced congruency effects, larger models achieved comparable performance across congruent and incongruent conditions, indicating that parameter count functions as an effective index of conflict resolution capacity. Most importantly, VLMs reproduced the fine-grained demand-resource relationship observed in human temporal dynamics \textbf{(Hypothesis 3)}: in high-conflict tasks, smaller models performed at chance on both trial types, whereas larger models showed above-chance performance on congruent trials but dropped below chance on incongruent trials. This pattern mirrors human conflict processing at short processing times and guards against spurious correlations \textbf{(Hypothesis 4)}: larger models specifically exhibit interference on incongruent high-conflict trials rather than monotonic improvement, demonstrating genuine conflict sensitivity rather than shortcut exploitation. Performance as a function of log parameter count approximated cross-sectional slices of human forced-response time-course functions, revealing a structural correspondence in how computational resources scale with conflict resolution demands. 

Notably, while this demand-resource relationship parallels human conflict processing, larger models' ceiling performance on high-conflict tasks does not fully recover to match low-conflict performance, unlike humans who eventually resolve conflict given sufficient PT. This suggests that current VLMs are limited by scale in zero-shot vision question answering, unable to achieve complete conflict resolution even at the largest model sizes tested. However, models demonstrably leveraging increased computational resources to resolve conflict indicates that the fundamental mechanisms of resource-dependent conflict resolution are operational.

These findings have profound implications for understanding the computational basis of cognitive control. Since the seminal work of \citet{cohen1990control}, numerous computational accounts of cognitive control have emerged, many using the Stroop task as a benchmark for modeling the congruency effect. In these models, processing pathways are explicitly modulated based on task demands: for instance, task demand units selectively enhance the color-naming pathway while suppressing the word-reading pathway, implementing a form of attention-based gating \citep{cohen1990control, verguts2017binding, kaplan2007modelling, herd2014neural, prabhakaran2025stroop}. Critically, many of these models—including \citet{cohen1990control}'s original parallel distributed processing (PDP) model—are trained via backpropagation, the same optimization method underlying modern deep learning. Yet they all incorporate hand-crafted architectural components specifically designed to implement specialized top-down control mechanisms. Contemporary VLMs, however, possess none of these specialized control architectures. They are trained end-to-end via standard gradient descent on next-token prediction, with no explicit task-switching mechanisms and no architectural components specifically designed for top-down control. Yet, as we have demonstrated, they exhibit the key signatures of human conflict processing. This dissociation suggests that cognitive control may not require specialized top-down modules, but rather can emerge naturally from today's deep learning architectures, with gradient-based learning allocating computational resources to resolve interference as model scale increases.

\vspace{-3mm}

\section{Conclusions and Future Works}

Cognitive control is widely recognized as fundamental to artificial general intelligence. Our findings demonstrate that human-like demand-resource relationships in conflict resolution emerge from optimization dynamics inherent in large-scale neural networks trained via standard gradient descent, without requiring specialized control architectures. This suggests that the adaptive flexibility needed to handle conflicting signals, noisy sensors, and adversarial inputs in real-world deployments may naturally emerge in sufficiently large models through scaling alone. 

Future work should pursue several critical directions. First, mechanistic interpretability studies are needed to isolate which architectural components and learned representations implement conflict detection and resolution, determining whether specific attention heads, layer types, or parameter regions specialize in control functions. Second, evaluations in naturalistic, in-the-wild settings beyond controlled laboratory tasks will reveal whether these emergent control capabilities generalize to the complex, ambiguous conflicts encountered in real-world deployments. Understanding these mechanisms will be essential for building robust AI systems capable of adaptive behavior under uncertainty.





\section*{Impact Statement}

This paper presents work whose goal is to advance the field of Machine
Learning. There are many potential societal consequences of our work, none
which we feel must be specifically highlighted here.

\bibliography{example_paper}
\bibliographystyle{icml2026}

\newpage
\appendix
\onecolumn

\section{Example Tasks}

\subsection{Examples of classic conflict tasks}

\vspace{-3mm}

\begin{figure}[H]
    \centering
    \includegraphics[width=1\linewidth]{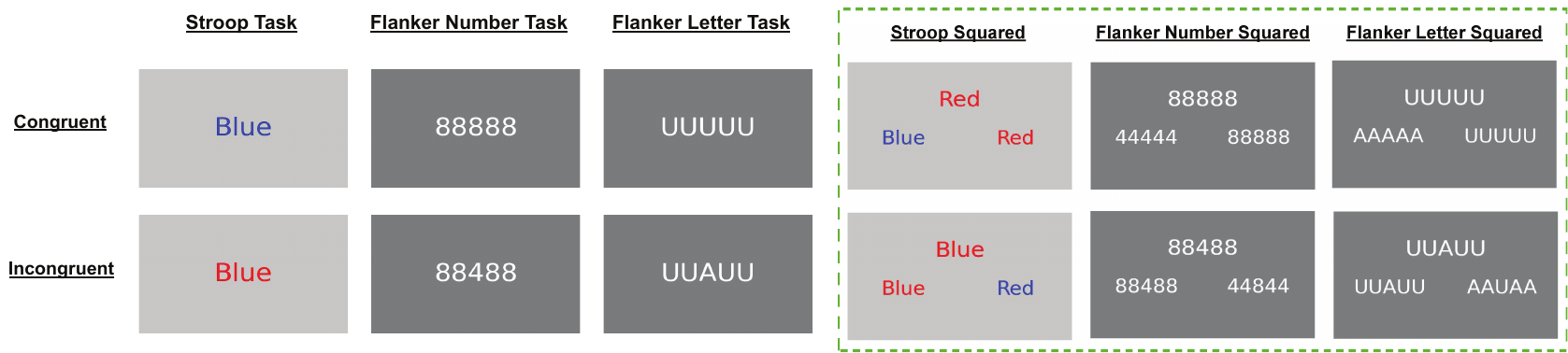}
    \caption{\textbf{Examples of classic conflict tasks: standard (top) and squared (bottom).} Standard tasks require models to ignore distractors (e.g., Stroop color vs. word, Flanker central vs. surrounding characters), while squared tasks require resolving dual mappings by selecting the option consistent with both target and distractor information, thereby increasing task complexity.}
    \label{fig:classic_conflict}
\end{figure}

\vspace{-5mm}

\subsection{Examples of object-level conflict tasks}
\begin{figure}[H]
    \centering
    \includegraphics[width=1\linewidth]{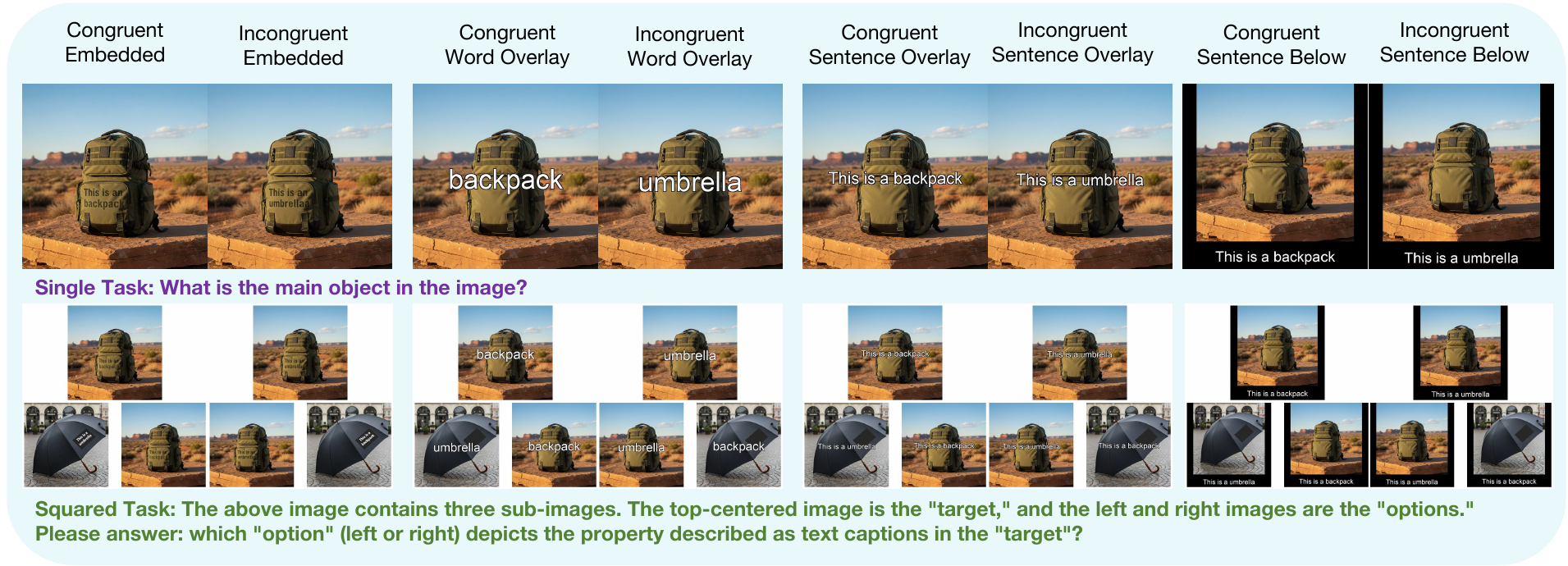}
    \caption{\textbf{Examples of object-level conflict tasks.} Images are paired with embedded captions that contradict the visual object identity (e.g., a backpack labeled as ``umbrella''). These tasks probe whether models can prioritize visual evidence over misleading textual information when recognizing object categories.}
    \label{fig:object_conflict}
\end{figure}

\vspace{-5mm}

\begin{figure}[H]
    \centering
    \includegraphics[width=1\linewidth]{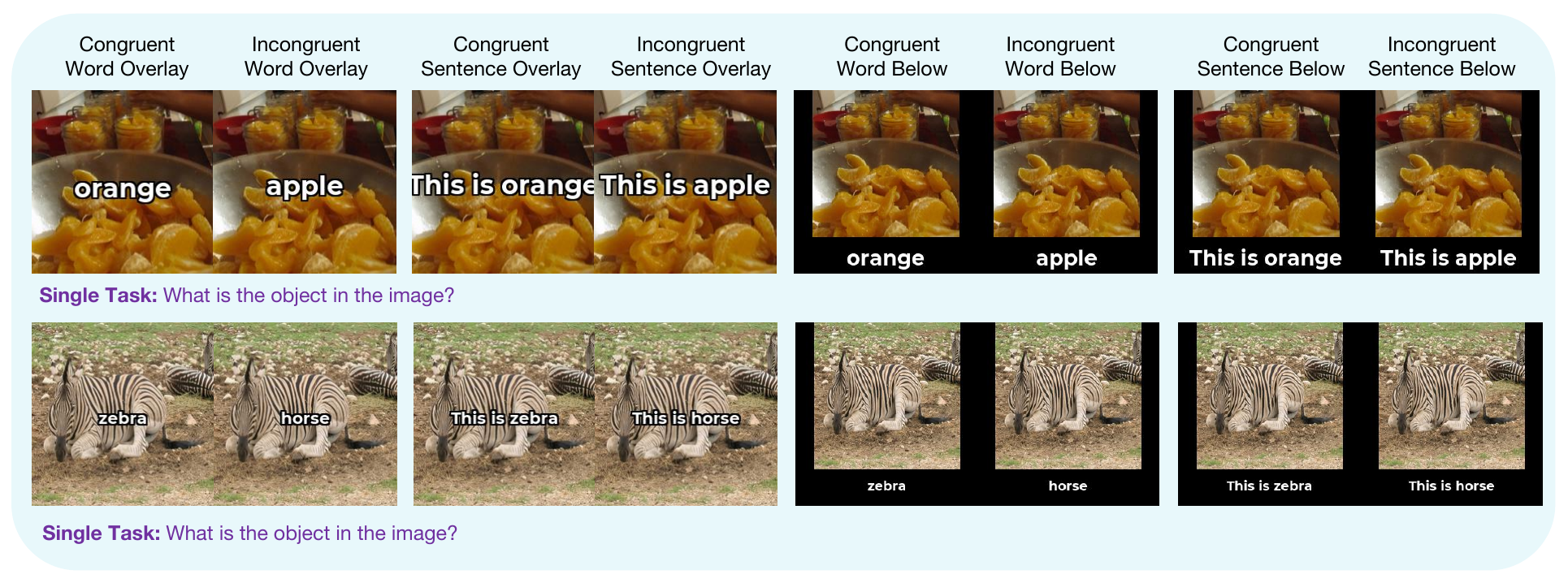}
    \caption{\textbf{Examples of object conflict tasks applied to natural images.} This figure illustrates varying text interference formats, including word overlay, sentence overlay, and captions placed below the image. The tasks test the model's robustness in correctly identifying visual objects (e.g., oranges, zebras) despite the presence of incongruent semantic labels (e.g., ``apple'', ``horse'').}
    \label{fig:coco_natural_conflict}
\end{figure}

\vspace{-5mm}

\subsection{Examples of property-level conflict tasks}
\begin{figure}[H]
    \centering
    \includegraphics[width=0.85\linewidth]{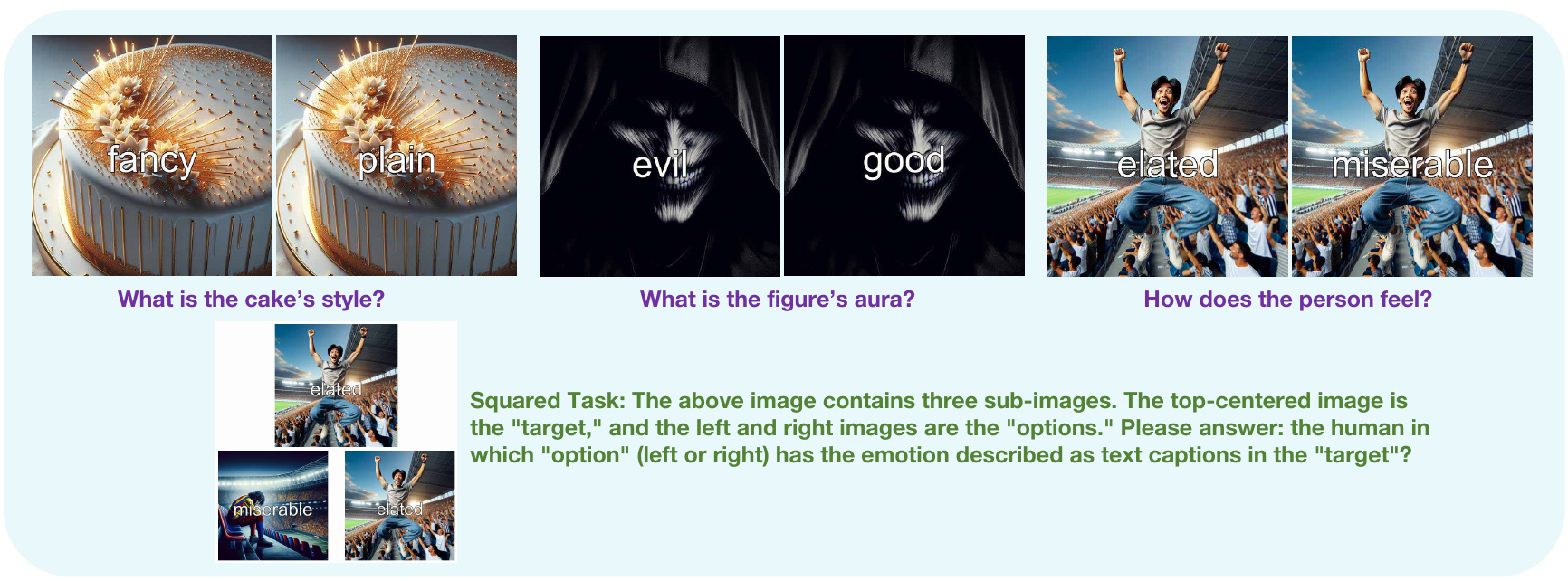}
    \caption{\textbf{Examples of property-level conflict tasks.} Unlike object identification, these tasks involve contradictions regarding specific visual attributes such as style (e.g., a fancy cake labeled as ``plain''), aura (e.g., an evil figure labeled as ``good''), or emotion (e.g., an elated person labeled as ``miserable''). The squared task variant (bottom) requires the model to correctly align the textual property with the corresponding visual option.}
    \label{fig:property_conflict}
\end{figure}

\vspace{-5mm}

\subsection{Examples of prior knowledge conflict tasks}
\begin{figure}[H]
    \centering
    \includegraphics[width=0.75\linewidth]{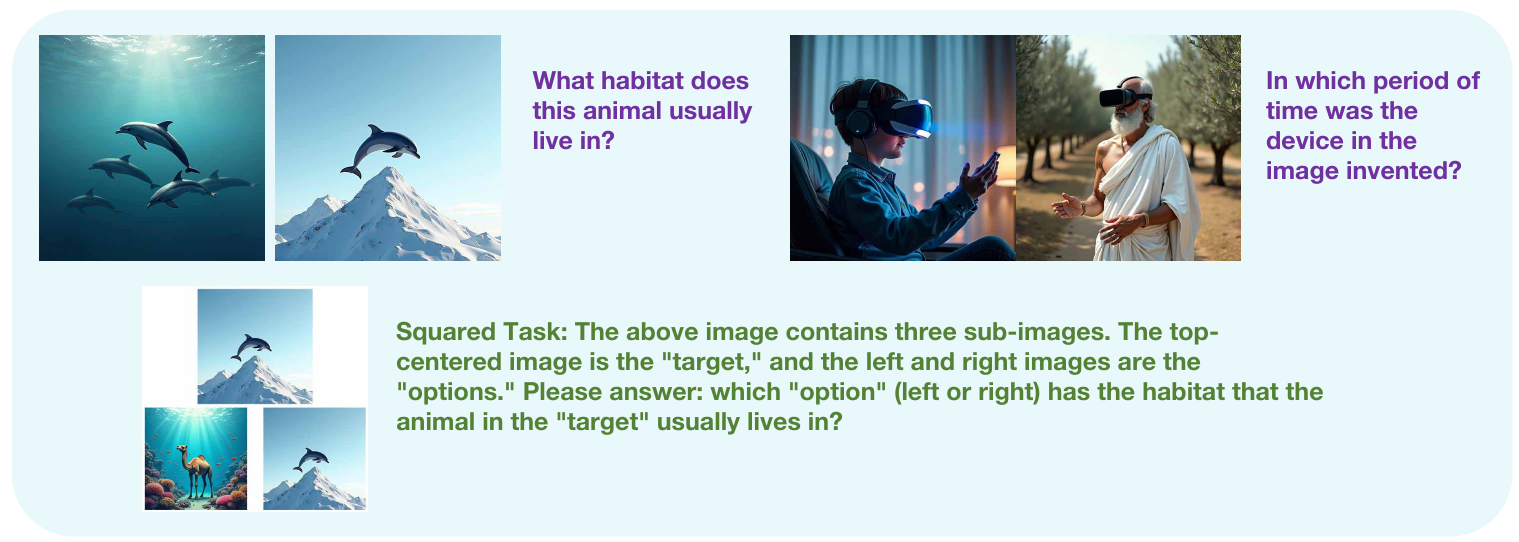}
    \caption{\textbf{Examples of prior knowledge conflict tasks.} In these scenarios, visual scenes contradict established semantic knowledge or common sense, such as animal habitats (e.g., a dolphin in the mountains) or historical timelines (e.g., VR headsets in ancient times).}
    \label{fig:prior_knowledge}
\end{figure}

\section{Backgrounds in Cognitive Science and AI Research}

\subsection{Processing Time As An Index of Computational Resources in Humans}
\label{app:processing_time}

In human psychophysics, reaction time (RT) is often treated as a proxy for the amount of \textbf{computational resources} deployed on a given task. However, the commonly used free-response paradigm mentioned above—in which participants are simply asked to respond as fast as possible-has long been recognized as a problematic index of underlying processing demands \citep{Pachella1974interpretation, Draheim2019reaction}. This is because such directions allow participants to complete each trial at their own discretion, so that the actual RT is being confounded by inter-individual differences in speed–accuracy trade-offs, general processing speed, and response strategies. Fast but error-prone participants may appear more efficient than slower but more accurate ones, and response latency often reflects motoric hesitation or habitual timing rather than the true time required for task-relevant computations \citep{Haith2016independence, Wong2017reaction}. As such, free RT does not reliably reflect \textbf{processing time (PT)}—the actual interval during which cognitive operations are being executed. Against this backdrop, recent work has reintroduced the forced-response paradigm into psychophysics research as a more principled alternative for estimating PT \citep{Dosher1976retrieval, Wickelgren1977speed}. In this method, response time is held constant across trials by requiring participants to respond exactly at a predetermined moment (e.g., in synchrony with a fixed "go" cue). PT is manipulated by varying the stimulus onset relative to this response cue, and the dependent measure is response accuracy as a function of available PT \citep{lee2025forced}. By decoupling the decision of when to respond from the process of deciding what to respond, the forced-response paradigm provides a clearer window into human performance in conflict tasks underlying limited computational resources.

\subsection{Symbolic and Connectionist Models}

A longstanding line of research in artificial intelligence has explored how control mechanisms can be embedded within integrated models of cognition. Prior to the rise of large-scale data-driven methods, general intelligence was often approached through symbolic and hybrid systems that explicitly specified internal structures for perception, memory, learning, and action selection—commonly known as cognitive architectures \citep{newell1989symbolic, anderson1983architecture, thagard2012cognitive}. Within these frameworks, cognitive control plays a central role in coordinating competing demands, prioritizing goals, and guiding adaptive behavior across tasks \citep{langley2009cognitive, salvucci2011toward}. Across a wide range of implementations, cognitive control consistently emerges as a core functional requirement for general-purpose intelligence \citep{kotseruba202040}. Architectures such as Soar and ACT-R offer contrasting yet complementary realizations of this faculty: Soar emphasizes hierarchical goal management and recursive subgoaling to resolve impasses \citep{young1999soar, laird2019soar}, while ACT-R employs production rules governed by sub-symbolic utility-based conflict resolution \citep{anderson1997act, ritter2019act}. Despite their differences, both reflect a shared commitment to modeling the internal regulation required for flexible, multi-context behavior \citep{stearns2020toward}. However, these architectures, built on human-designed knowledge and representations, often struggle to scale and generalize compared to those that rely on general-purpose learning and search algorithms leveraging computational resources \citep{sutton2019bitter}.

In parallel, modern connectionist approaches in deep learning have pursued similar regulatory functions through distributed and data-driven architectures. While these models have achieved notable successes in perception and pattern recognition, they continue to face limitations in core aspects of cognitive control—particularly in generalizing across tasks, applying learned rules to novel situations, and flexibly adapting to shifting goals and environments \citep{russin2020deep}. These limitations closely mirror challenges addressed by the prefrontal cortex (PFC), which plays a central role in rule-based behavior, abstract reasoning, and dynamic coordination \citep{miller2000prefontral}. Neuroscientific evidence implicates the PFC in managing working memory, resolving interference, and exerting top-down modulation over other brain areas—functions that remain underdeveloped in current AI systems \citep{smith1999storage, miller2001integrative}. Efforts to bridge this gap explore architectural principles that integrate connectionist learning with mechanisms of cognitive control, offering promising directions for developing more generalizable artificial agents. Recent theoretical frameworks have proposed that the computational processes underlying human cognitive control may be grounded in associative learning \citep{abrahamse2016grounding}. However, this perspective still awaits empirical support, particularly from large-scale, general-purpose neural networks that are subjected to scaling.

\section{Pipeline for AI-Generated Images}
\label{appen_pipeline_for_aigen}

\begin{figure}[ht]
    \centering
    \includegraphics[width=0.65\linewidth]{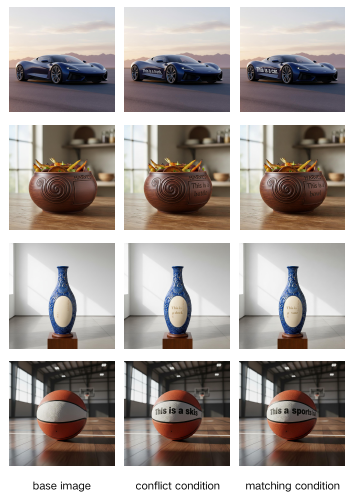}
    \caption{AI-generated image examples produced by our three-stage pipeline. From left to right: base image of the target object, conflict condition with a naturalistic overlay declaring ``This is a \texttt{\{conflict\_class\}}'', and matching condition where only the overlaid text is replaced with ``This is a \texttt{\{original\_class\}}'' while preserving style and placement.}
    \label{fig:nanobanana examples}
\end{figure}

We generate image triplets using Google's multimodal generative model Gemini-2.5-flash-image-preview \cite{}. For each object class drawn from a COCO \cite{}, we first synthesize a photorealistic base image of the target object in a clean scene with space suitable for text integration. We then apply two in-model editing passes: (i) seamlessly overlay a short declarative phrase labeling the object as a conflict class to produce a contradiction image; and (ii) starting from that edit, replace only the overlaid text with the original class name while preserving its position, font, color, and style to obtain a matching image. The prompts used are listed below.
Class pairings follow a simple sampling strategy (same-supercategory by default) and the pipeline is repeated to produce multiple sets for analysis. All synthesis and edits are performed by the model (no local inpainting or text rendering). 
Some examples are shown in Fig~\ref{fig:nanobanana examples}.

\textbf{Prompt templates for the three-stage triplet-generation pipeline:}

(1) base image synthesis for \texttt{\{original\_class\}}: 

A photorealistic image of a detailed \texttt{\{original\_class\}} in a realistic setting. The object is in clear focus and well-lit. There are no distracting elements in the background. Try to make the generated image have enough area to display text on the main object (such as the car body).

(2) conflict overlay declaring ``This is a \texttt{\{conflict\_class\}}'':

Find the biggest empty area in the body of the main item in the image and then naturally embed the text ``This is a \texttt{\{conflict\_class\}}'' to generate the final image. Be careful to use clever embedding techniques to ensure the text fits naturally. For example, for a horse image, you can attach it to the horse's body. For a bus image, you can remove the information on the bus body and then embed the new text. The embedded text should be in a color that's clearly visible and cover the original image information.

(3) matching replacement to ``This is a \texttt{\{original\_class\}}'':

In this image, replace the text ``\texttt{\{conflict\_class\}}'' with ``\texttt{\{original\_class\}}'' while keeping exactly the same text style, font, color, size, and location. Do not change anything else about the image - only replace the text content. Make sure the new text ``\texttt{\{original\_class\}}'' appears in the exact same position and style as the original text ``\texttt{\{conflict\_class\}}''.


\section{Results By Task Types}

\begin{table}[H]
\caption{Mean accuracies across task types and designs. *$p<0.05$, ***$p<0.001$ (paired $t$-tests).}
\label{tab:combined-accuracies}
\setlength{\tabcolsep}{5pt}
\centering
\begin{tabular}{llcccc}
\toprule
\textbf{Task Type} & \textbf{Design} & \textbf{Cong.} & \textbf{Incong.} & \textbf{Diff.} & \textbf{p-value} \\
\midrule
\multicolumn{6}{l}{\textit{Non-Squared Tasks}} \\
\midrule
Classic Conflict & Stroop-content & 0.953 & 0.938 & 0.015 & 6.64e-01 \\
Classic Conflict & Stroop-color & 0.968 & 0.931 & 0.037 & 1.53e-01 \\
Classic Conflict & Flanker-letter & 0.873 & 0.860 & 0.013 & 3.17e-01 \\
Classic Conflict & Flanker-number & 0.859 & 0.824 & 0.035* & 1.47e-02 \\
Object-level Conflict & Embedded & 0.864 & 0.851 & 0.013 & 3.37e-01 \\
Object-level Conflict & Word Overlay & 0.977 & 0.813 & 0.164*** & 5.32e-13 \\
Object-level Conflict & Sentence Overlay & 0.977 & 0.825 & 0.152*** & 2.11e-12 \\
Object-level Conflict & Sentence Below & 0.976 & 0.810 & 0.166*** & 1.29e-10 \\
Property-level Conflict & -- & 0.971 & 0.562 & 0.409*** & 7.01e-20 \\
Knowledge Prior Conflict & -- & 0.951 & 0.694 & 0.258*** & 1.82e-07 \\
\addlinespace[3pt]
\midrule
\multicolumn{6}{l}{\textit{Squared Tasks}} \\
\midrule
Classic Conflict & Stroop & 0.591 & 0.366 & 0.225*** & 1.30e-06 \\
Classic Conflict & Flanker-number & 0.578 & 0.344 & 0.234*** & 1.92e-06 \\
Classic Conflict & Flanker-letter & 0.564 & 0.329 & 0.235*** & 3.58e-06 \\
Object-level Conflict & Embedded & 0.643 & 0.372 & 0.271*** & 6.09e-05 \\
Object-level Conflict & Word Overlay & 0.719 & 0.265 & 0.454*** & 3.40e-09 \\
Object-level Conflict & Sentence Overlay & 0.703 & 0.280 & 0.423*** & 1.37e-08 \\
Object-level Conflict & Sentence Under & 0.718 & 0.276 & 0.443*** & 8.99e-09 \\
Property-level Conflict & -- & 0.737 & 0.337 & 0.400*** & 2.31e-09 \\
Knowledge Prior Conflict & -- & 0.682 & 0.499 & 0.184* & 1.31e-02 \\
\bottomrule
\end{tabular}
\end{table}

\newpage

\section{Results by Models (Top 30)}

\begin{figure}[h]
    \centering
    \includegraphics[width=1.0\linewidth]{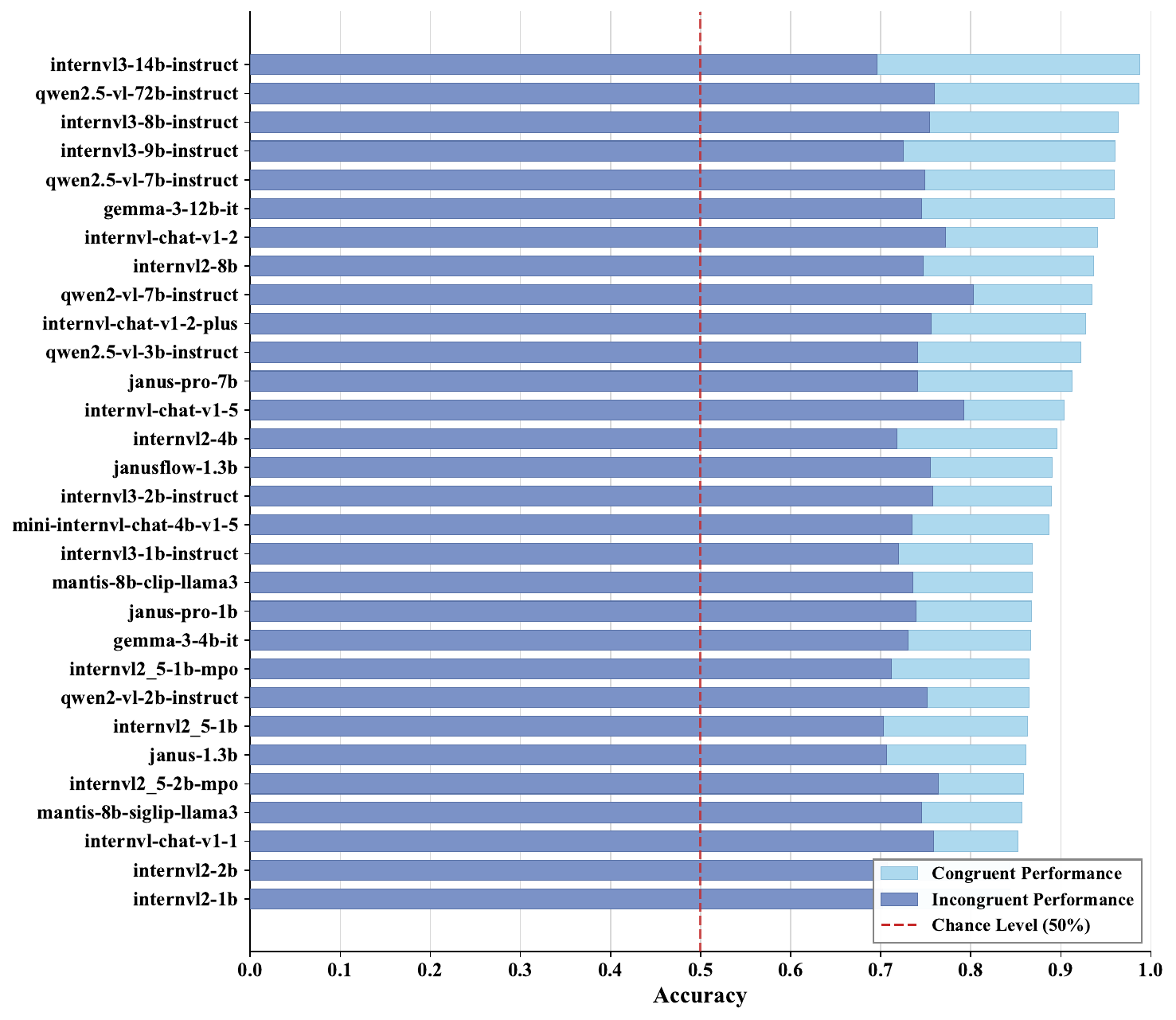}
    \caption{Top 30 Models: Congruent vs Incongruent (Non-Squared)}
    \label{fig:non-squared}
\end{figure}

\begin{figure}[h]
    \centering
    \includegraphics[width=1.0\linewidth]{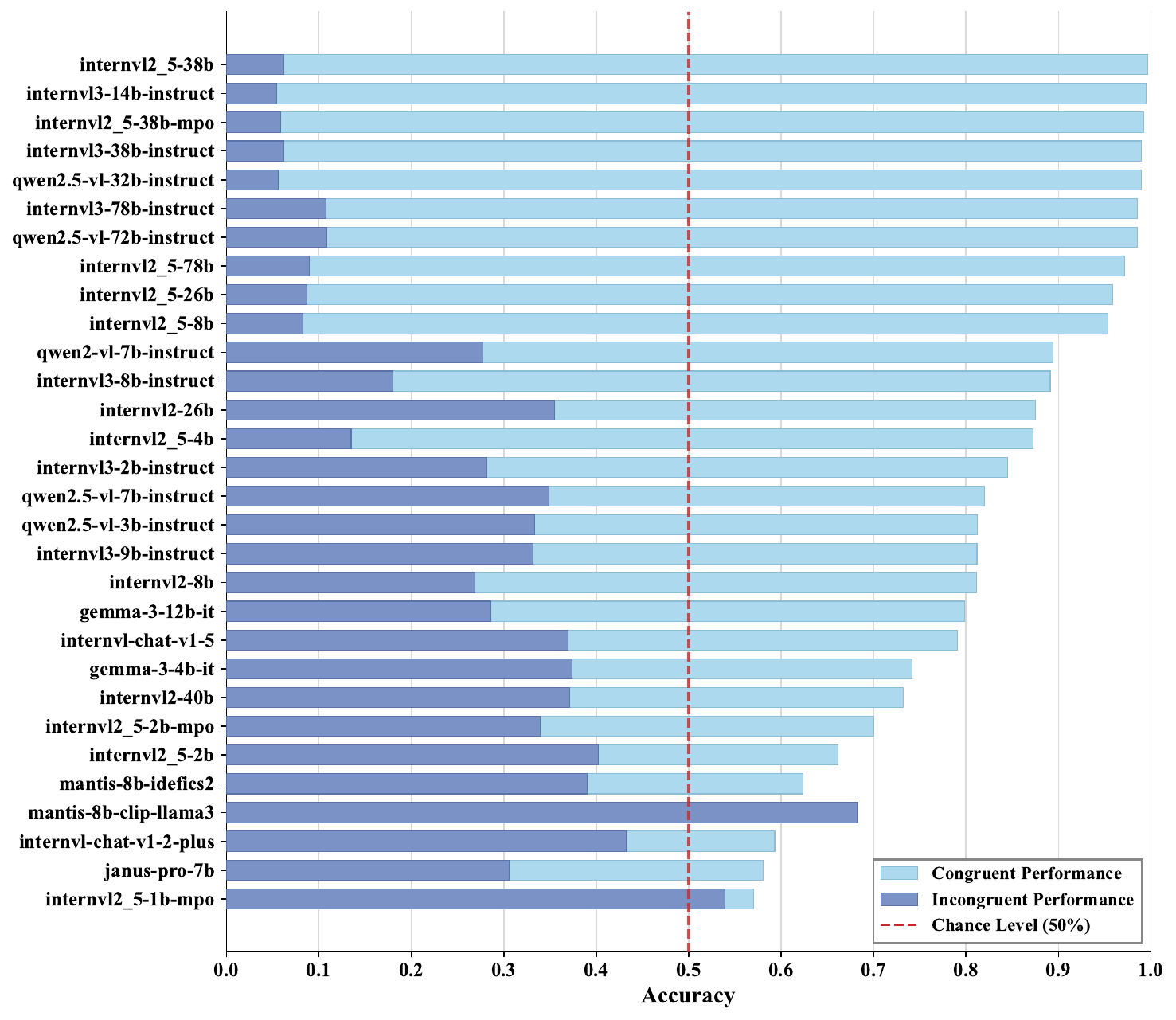}
    \caption{Top 30 Models: Congruent vs Incongruent (Squared)}
    \label{fig:squared}
\end{figure}

\end{document}